\documentclass{article}

\PassOptionsToPackage{numbers}{natbib}
\usepackage[preprint]{neurips_2026}

\usepackage[utf8]{inputenc}
\usepackage[T1]{fontenc}
\usepackage{hyperref}
\usepackage{url}
\usepackage{booktabs}
\usepackage{amsfonts}
\usepackage{amsmath}
\usepackage{nicefrac}
\usepackage{microtype}
\usepackage{xcolor}
\usepackage{graphicx}
\usepackage{multirow}
\usepackage{makecell}
\usepackage{pifont}
\usepackage{enumitem}
\usepackage{algorithm}
\usepackage{algpseudocode}
\usepackage{xcolor}
\usepackage[normalem]{ulem}

\DeclareRobustCommand{\sysname}{SimWeaver}
\DeclareRobustCommand{\sysasset}{\sysname\nobreakdash-Asset}
\DeclareRobustCommand{\syssim}{\sysname\nobreakdash-Sim}
\DeclareRobustCommand{\syssyn}{\sysname\nobreakdash-Syn}
\DeclareRobustCommand{\sysreal}{\sysname\nobreakdash-Real}

\title{\sysname{}: Zero-Shot RGB Sim-to-Real for Deformable Manipulation}

\author{%
  \textbf{Wenkang Hu\textsuperscript{1}\quad Haoran Wang\textsuperscript{1}\quad Yitong Li\textsuperscript{1}\quad Liu Liu\textsuperscript{2}\quad Mengao Zhao\textsuperscript{2}\quad Lai Jiang\textsuperscript{1}\quad Xincheng Tang\textsuperscript{1}} \\
  \textbf{Junhang Wei\textsuperscript{3}\quad Zhengjie Shu\textsuperscript{1}\quad Zhendong Wang\textsuperscript{3}\quad Zhizhong Su\textsuperscript{2}\quad Huamin Wang\textsuperscript{3}\quad Ruigang Yang\textsuperscript{1,\,$\dagger$}} \\
  \vspace{3pt} \\
  \normalfont
  \textsuperscript{1}Shanghai Jiao Tong University\quad
  \textsuperscript{2}Horizon Robotics\quad
  \textsuperscript{3}Style3D Research \\
  \texttt{\small $\dagger$~Corresponding author}
}

\begin{document}
\maketitle

\begin{abstract}
RGB sim-to-real for deformable manipulation has remained largely unsolved without real-world fine-tuning. We present \textsc{\sysname}, which trains zero-shot RGB VLA policies on 200 simulated demonstrations per task, reaching above $80\%$ per-task and $91\%$ average real-world success across 5 diverse deformable tasks including plastic-bag manipulation, without teleoperation or per-task calibration. \sysname{} combines a reliable measurement-backed simulator (\syssim) with an extensible asset framework supporting single-image generation (\sysasset), a deterministic topology-aware trajectory synthesizer (\syssyn), and a sim-to-real protocol with ISP-aware photometric augmentation (\sysreal). On silk grasping, the sim-trained policy reaches $100\%$ under visual distribution shifts where real-data baselines drop to $9$--$70\%$, at two orders of magnitude lower per-trajectory cost. We will release \sysname{} and a representative asset subset. Project page: \url{https://simweaver.github.io/}
\end{abstract}

\section{Introduction}
\label{sec:intro}

\begin{figure}[!t]
  \centering
  \includegraphics[width=\linewidth]{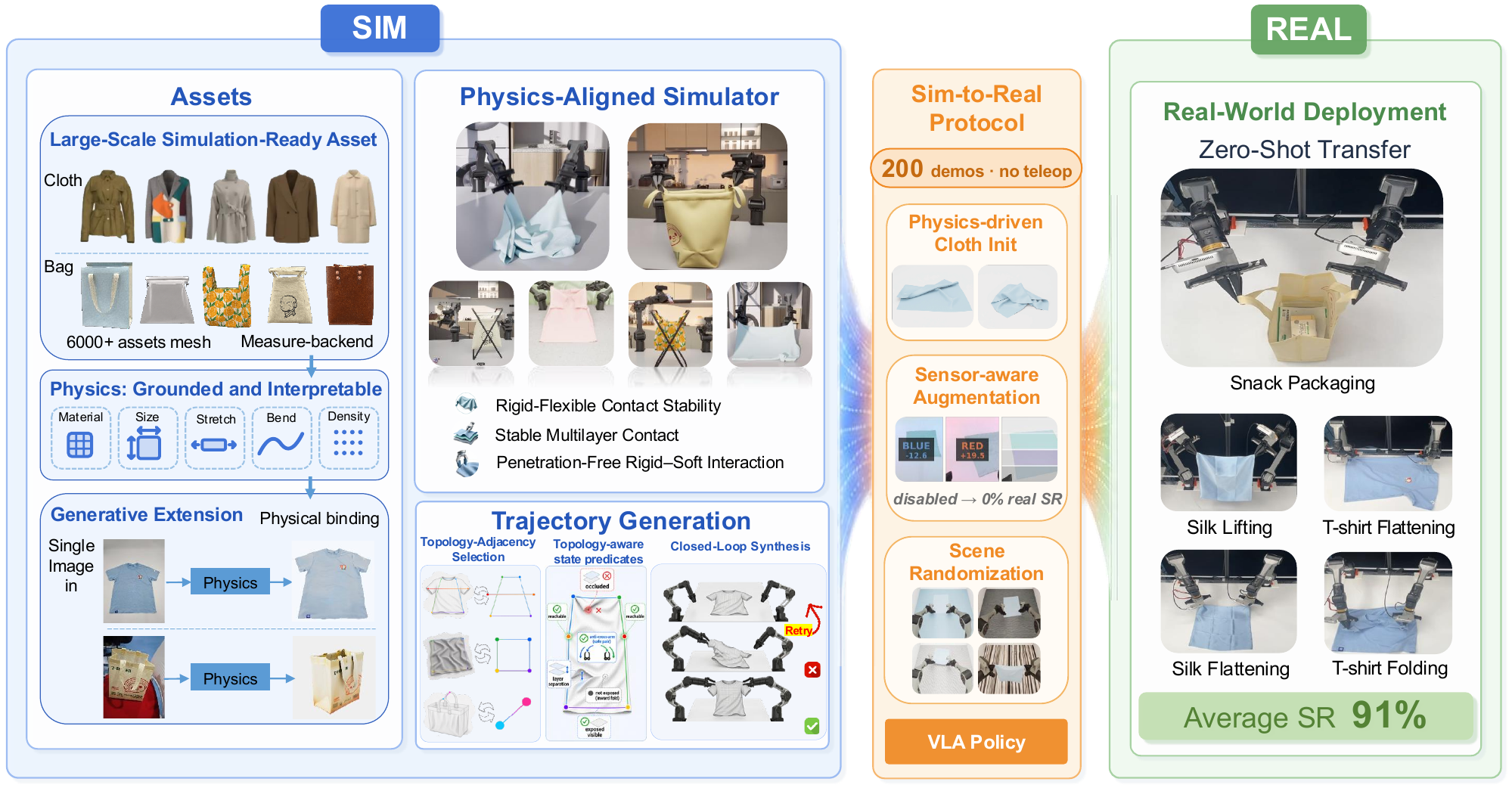}
  \caption{\textbf{System architecture.} Assets feed the simulator; \protect\syssyn{} produces 200 deterministic demonstrations per task from a single seed; the demonstrations train an end-to-end VLA policy that is deployed zero-shot on real hardware. The pipeline also supports closed-loop policy evaluation in simulation.}
  \label{fig:overview}
\end{figure}

Recent studies~\citep{lin2024data_scaling,bu2025agibot} show that policy learning for robotic manipulation follows an approximate power-law relationship with respect to the number of training demonstrations and environments. Real-world data offer the highest fidelity for policy learning but are labor-intensive, costly, and difficult to scale across environments. While such collection is tractable for rigid-object manipulation, it becomes prohibitive for deformable objects due to large configuration spaces, non-linear shape variation, and complex interactions with rigid end-effectors. To enable scalable data generation, the community has extensively explored synthetic demonstrations via physics-based simulation. Although effective for rigid manipulation, prior approaches have not achieved success in deformable-object settings without real-world fine-tuning. 

Several factors contribute to this limitation. Widely used simulators, such as NVIDIA Isaac Sim, exhibit unreliable contact dynamics for thin structures and fail to capture realistic physical behavior of deformable materials. Consequently, demonstration synthesis pipelines developed for rigid-body manipulation transfer poorly to deformable settings. Moreover, pixel-based end-to-end policy learning for deformable object manipulation has stagnated due to the unresolved sim-to-real gap, prompting a shift in the community toward depth and point-cloud representations over raw visual inputs.

Concurrent SIM1~\citep{zhou2026sim1} demonstrates pixel-based sim-to-real transfer for garment folding, achieving 87\% in-domain real-world success on T-shirt folding. However, this relies on per-task expert-guided material calibration and teleoperated demonstrations for trajectory generation. These constraints leave unresolved whether pixel-based sim-to-real transfer for deformable manipulation can be made teleoperation-free, scalable beyond folding-centric tasks, and deployable without per-asset manual calibration.

In this work, we present \emph{\sysname}, a sim-to-real framework that addresses these three constraints (Fig.~\ref{fig:overview}). \sysname{} integrates a deformable simulator (\syssim) with an extensible asset framework (\sysasset), a topology-aware trajectory synthesizer (\syssyn), and a sim-to-real deployment protocol (\sysreal). Unlike point-cloud-based pipelines, \sysname{} adopts a pixel-based formulation that robustly handles visually challenging materials including dark, low-texture, and reflective fabrics.

We summarize our main contributions as follows:
\begin{itemize}[leftmargin=1.5em, itemsep=2pt, topsep=2pt]
 \item \textbf{Zero-shot RGB sim-to-real for deformable 
    manipulation with only 200 synthesized demonstrations per task.}
    \sysname{} achieves an average of 91\% real-world success 
    ($>$80\% per task) across 5 diverse deformable tasks—including 
    plastic-bag manipulation—without teleoperation or per-task 
    calibration. On silk grasping, the sim-trained policy maintains 
    100\% under visual distribution shifts where real-robot-trained
    baselines drop to 9--70\%.                                                            
  
  \item \textbf{Reliable deformable simulator with robust collision handling, measurable physical correspondence, and an extensible asset framework.} \sysname-Sim introduces robust collision handling, 
    penetration-prevention mechanisms, and trajectory-replay 
    determinism, addressing contact-reliability failures of 
    widely-used solvers (Isaac Sim, VBD-based) on thin fabrics, 
    with physical parameters in direct real-world correspondence. 
    The simulator additionally supports an extensible asset framework 
    (\sysname-Asset) for mesh import and single-image asset 
    generation across diverse deformable categories including 
    plastic bags with handle structures.                 
  
\item \textbf{Topology-aware trajectory synthesis without 
    teleoperation.}
    \sysname-Syn generates high-quality deterministic demonstrations 
    from a single seed without learned generative models, 
    teleoperation, or post-hoc discriminator filtering.                               
  \end{itemize}                                                                                                                         

\section{Related Work}
\label{sec:related}

\subsection{Deformable Simulation}

General-purpose deformable simulators such as SoftGym~\citep{lin2020softgym},
DiffCloth~\citep{li2022diffcloth}, PlasticineLab~\citep{huang2021plasticinelab},
and DaXBench~\citep{chen2023daxbench} cover cloth, rope, fluids, and
elastoplastic materials, but are not primarily designed for
robot--deformable interaction and often suffer from unreliable grasping
contact, instability in contact-rich regimes, or limited support for
thin-shell objects with handles.

Recent robot-oriented simulators reduce this gap. GarmentLab~\citep{lu2024garmentlab} and DexGarmentLab~\citep{wan2025dexgarmentlab} support large-scale policy learning for garment manipulation with Position-Based Dynamics~\citep{muller2007position} for cloth simulation, but remain tailored to standard fabrics and depend on simulator-specific contact tuning rather than physically measurable parameters. Concurrent VBD-based~\citep{chen2024vbd} solvers including SIM1~\citep{zhou2026sim1} further improve fidelity for robot--deformable interaction; remaining failure modes (cloth--rigid penetration, grasp instability, replay non-determinism) are characterized in \S\ref{sec:sim}.

\subsection{Sim-to-Real for Deformable Manipulation}

Deformable sim-to-real has split along observation modality. Point-cloud methods~\citep{ha2021flingbot, weng2021fabricflownet, canberk2022clothfunnels, xue2023unifolding, sundaresan2022diffcloud, lu2024garmentlab, wan2025dexgarmentlab} achieve effective transfer on standard fabrics by leveraging explicit geometry, but become unreliable when depth is noisy or incomplete---dark, reflective, low-texture, or self-occluded materials. RGB end-to-end has historically lagged: VSF~\citep{hoque2020vsf} reported pure RGB underperforming RGBD by 80\% on fabric folding, and the gap has not closed---DexGarmentLab's RGB diffusion policy reaches only a 58\% mean zero-shot real success across four garment tasks~\citep{wan2025dexgarmentlab}, and recent RGB diffusion policies on bimanual deformables continue to struggle. Concurrent work SIM1~\citep{zhou2026sim1} shows RGB sim-to-real transfer for garment folding, but relies on teleoperated trajectories and post-hoc filtering (head-to-head in \S\ref{sec:exp-gen}).

Plastic-bag manipulation is largely real-robot-first: AutoBag relies on visual markers~\citep{chen2023autobag}, ShakingBot studies dynamic shaking on a specific archetype~\citep{chen2023shakingbot}, DextAIRity uses a multi-arm airflow setup on one bag~\citep{xu2022dextairity}, and SOI learns dynamics from real point clouds due to unreliable bag simulation~\citep{zhou2024soi_bag}. Simulation benchmarks remain simplified---DeformableRavens covers single-topology bags~\citep{seita2021bags}, SoftMimicGen omits handle structure~\citep{moghani2026softmimicgen}---because thin-shell deformation, 2-torus handles, self-contact, and large pick-and-place deformation resist standard tuning. Our pipeline handles bags with explicit handles within the same RGB sim-to-real protocol as garment tasks (\S\ref{sec:sim}, \S\ref{sec:exp}).

\subsection{Trajectory Generation for Manipulation}
Trajectory generation expands a small set of demonstrations into larger training datasets. Existing methods are most developed for rigid manipulation: MimicGen~\citep{mandlekar2023mimicgen}, SkillMimicGen~\citep{garrett2024skillmimicgen}, and DexMimicGen~\citep{jiang2024dexmimicgen} synthesize demonstrations by transforming source trajectories through object-centric rigid-frame correspondence. This assumption breaks down for deformables, which lack canonical pose frames and undergo large non-rigid (sometimes topological) deformation.

Deformable extensions use non-rigid registration, task-specific correspondence, or learned generation. SoftMimicGen~\citep{moghani2026softmimicgen} warps trajectories via non-rigid registration but stays at 17\% on bimanual towels even at 750 generated demos, partly because warping enforces no IK feasibility or bimanual consistency. Garment-specific methods exploit canonical structure---DexGarmentLab/HALO~\citep{wan2025dexgarmentlab} via sleeve/collar correspondence, FoldNet's KG-DAgger~\citep{chen2025foldnet} via keypoint-driven recovery---but neither transfers to non-garment deformables such as bags. Concurrent SIM1~\citep{zhou2026sim1} pairs diffusion-based generation with post-hoc filtering at the cost of teleoperation seeds.

These methods follow a generate-then-filter paradigm. In contrast, our generator is deterministic and topology-aware, producing feasible high-quality trajectories without learned generative models, teleoperation seeds, or post-hoc rejection (\S\ref{sec:gen}, \S\ref{sec:exp}).

\section{\syssim{}: Reliable Deformable Simulation}
\label{sec:sim}
Leveraging large-scale low-cost simulation data is promising for robot policy learning, especially for VLA training. However, the sim-to-real gap degrades real-world performance of policies trained in simulation. This challenge is particularly severe in deformable object manipulation, where rigid-flexible interactions among deformable objects, robots, and the environment introduce substantial physical complexity. Most methods therefore combine simulation pretraining with limited real-world adaptation. Since sim-to-real performance fundamentally depends on simulator fidelity, improving simulator realism can substantially enhance zero-shot real-world performance~\citep{zhou2026sim1}.

\subsection{Base Simulator}

The choice of the simulator is critical for reducing the sim-to-real gap in synthetic demonstration generation. We identify three key requirements for an effective deformable manipulation simulator:

\textbf{Fabric realism:} stable simulation of diverse deformable materials with varying physical properties, ranging from soft to stiff and thin to thick;
\textbf{Multilayer contact stability:} robust handling of contacts among multilayer fabrics and multiple deformable objects;
\textbf{Rigid-flexible contact stability:} reliable simulation of interactions between rigid robot end-effectors and deformable objects, particularly during grasping and compression.

To select the simulation backbone of \syssim{}, we investigate three families of cloth simulators commonly used in the community: Position-Based Dynamics (PBD)~\citep{muller2007position} as integrated in NVIDIA Isaac Sim, a VBD-based simulator~\citep{chen2024vbd}, and the RGBench simulator~\citep{hu2026RGBench}. We compare them along the three requirements above.

We adopt the RGBench simulator~\citep{hu2026RGBench} (our prior work) as \syssim{}'s base solver. Rather than proposing a new solver, our contribution is to make it contact-reliable for closed-loop robot manipulation: we eliminate the penetration, grasp-contact jitter, and run-to-run replay non-determinism that otherwise persist under contact-rich grasping (\S\ref{sec:sim-design}).

\begin{figure}[t]
  \centering
  \begin{minipage}[b]{0.66\linewidth}
    \centering
    \includegraphics[width=\linewidth]{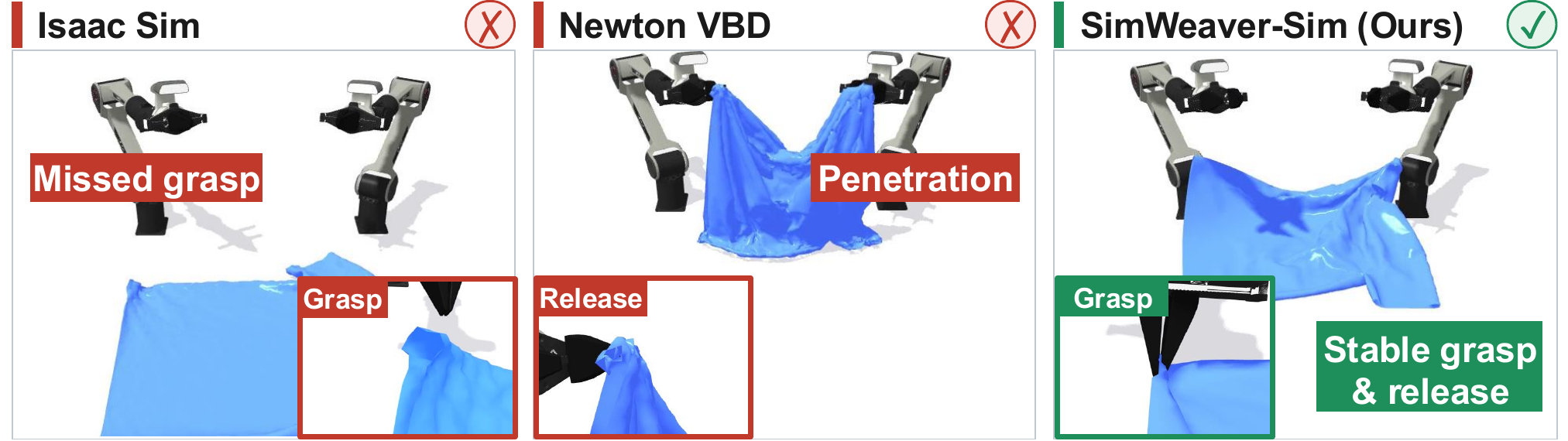}\\[2pt]
    {\footnotesize (a) Isaac Sim (PBD) \quad (b) VBD-based pipeline \quad (c) \syssim{} (ours)}
  \end{minipage}\hfill
  \begin{minipage}[b]{0.32\linewidth}
    \centering
    \includegraphics[width=\linewidth]{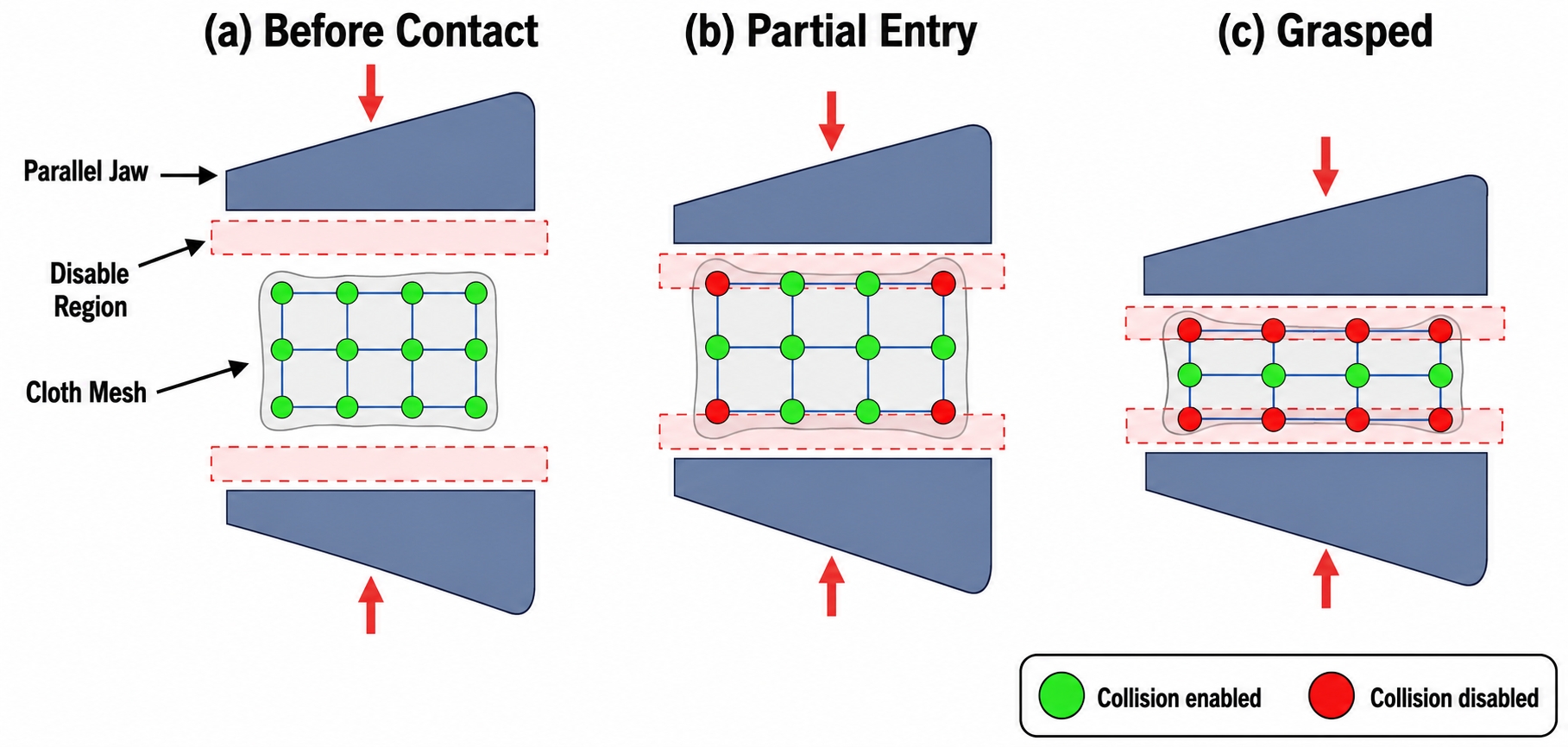}\\[2pt]
    {\footnotesize (d) Active collision region}
  \end{minipage}
  \caption{\textbf{Simulator failure modes (a--c) and \syssim{}'s active-collision-region scheme (d).} (a) Isaac Sim with PBD cloth exhibits instability and fails to grasp the garment. (b) Newton VBD produces cloth--rigid penetration during dual-arm manipulation. (c) \syssim{} achieves stable, penetration-free behavior under identical trajectories. (d) The green region denotes the active collision zone for cloth self-collision handling, while the surrounding collision-forbidden margin suppresses grasping jitter.}
  \label{fig:sim-failures}
\end{figure}

As illustrated in~\autoref{fig:sim-failures}, (a) NVIDIA Isaac Sim exhibits instability and fails to grasp the garment; (b) Newton VBD produces visible robot--cloth penetration during dual-arm manipulation. In contrast, (c) \syssim{} reproduces realistic T-shirt dynamics with stable multilayer contact behavior closely matching real-world observations and remains penetration-free under contact-rich grasping, owing to the active-collision-region scheme detailed in \S\ref{sec:sim-design}.


\subsection{Optimized Robot-Deformable Interaction}
\label{sec:sim-design}

Contact reliability---particularly robot--deformable interaction involving thin and visually challenging fabrics---remains a major obstacle to sim-to-real transfer in deformable manipulation. Simulated trajectories that contain penetration artifacts or unstable contacts are typically infeasible to execute on real hardware.

Multiple factors contribute to this challenge. Fundamentally, the absence of tactile feedback in robotic grippers causes both simulated and real-world jaws to close with near-maximal force to prevent slippage, leaving minimal clearance for multilayer fabrics. Accurately resolving such interactions in simulation is difficult: it requires precise force-aware contact control and realistic friction modeling between rigid and deformable bodies.

To mitigate this issue without compromising demonstration quality, we introduce a simple but effective strategy for stabilizing robot--deformable interactions (\autoref{fig:sim-failures}(d)). We define a collision-forbidden region around each gripper jaw, within which deformable self-collision handling is disabled, and adaptively compute a complementary \emph{active collision region} in which cloth self-collisions remain enabled, requiring that all intersections be resolved through the simulator's built-in untangling mechanism before particles leave this region. When the gripper closes and the jaw clearance becomes small, the active region contracts, suppressing jitter and oscillation; when the gripper opens, the region expands to provide sufficient space for stable intersection resolution. Owing to the efficiency of the underlying solver, this strategy introduces no observable artifacts in the synthesized demonstrations and no measurable degradation in sim-to-real transfer fidelity.


\subsection{Quantitative Reliability Comparison}
\label{sec:sim-comparison}

To validate that our RGBench-based simulator substantially reduces the sim-to-real gap for zero-shot deformable manipulation, we evaluate its contact reliability and physical realism across three standardized scenarios: thin cloth–rigid interaction, cloth grasping, and garment folding.

For each setting, we report quantitative metrics over $n$ trials per condition in Section~\ref{sec:exp-sim}. \syssim{} achieves significantly lower penetration rates and higher trajectory success rates than both NVIDIA Isaac Sim cloth simulation and a VBD-based baseline, while maintaining competitive simulation efficiency. These reliability improvements directly enable the downstream sim-to-real performance reported in Section~\ref{sec:exp}.



\section{\syssyn{}: Automatic Topology-Aware Trajectory Synthesis}
\label{sec:gen}

\syssyn{} synthesizes training trajectories without teleoperated demonstrations. Existing learned samplers~\citep{zhou2026sim1} still rely on human teleoperation to seed generative models, degrade under shifts in the cloth's initial pose, and report pass rates below 50\% on bimanual cloth tasks. Our pipeline removes this dependency through three deterministic components: \textbf{topology-adjacency selection} (\textsc{tas}) defines semantically valid bimanual grasp pairs from the asset's canonical mesh; \textbf{topology-aware state predicates} determine which pairs are feasible under the current observation; and \textbf{closed-loop synthesis} verifies each grasp attempt and reselects when execution fails. A constrained planner then realizes the selected pairs as smooth, executable trajectories (Appendix~\ref{app:planner-ik-backends}); diversity comes independently from randomized simulator initializations.

\paragraph{Topology-adjacency selection (\textsc{tas}).}
Across deformable objects, manipulation semantics are determined not by keypoints alone but by their topological relations: the same pair of corners may unfold or fold a cloth depending on which two are grasped. We represent feasible bimanual configurations as a labeled graph $G=(V,E)$ on the canonical mesh, where nodes $V$ are semantic landmarks (cloth corners, T-shirt neck/hem/shoulder/sleeve points, bag strap tips) and edges $(u,v)\in E$ are task-feasible grasp pairs. $G$ is extracted automatically for symmetric assets and labeled once per asset class otherwise (Appendix~\ref{app:topo-instantiations}). Given an observation $\text{obs}_t$, we select
\begin{equation}
(l^{\star}, r^{\star}) \;=\; \arg\max_{(u,v)\,\in\,E\,\cap\,\mathcal{F}(\text{obs}_t)}\; S(u,\,v\,;\,\text{obs}_t),
\label{eq:select}
\end{equation}
where $\mathcal{F}(\text{obs}_t)$ is the geometric feasibility set on the current cloth state and $S$ scores the surviving candidates. The adjacency graph $G$ does most of the work: by restricting the argmax to $E$, all non-adjacent pairs are excluded by construction and never enter the score. $S$ is then a lightweight, task-specific selector among feasible adjacency edges. Unlike continuous-registration methods that match deformation fields point by point~\citep{moghani2026softmimicgen}, our discrete graph formulation applies uniformly across cloth, garment, and bag topologies; only $G$ changes across assets.

\paragraph{Topology-aware state predicates.}
A topologically valid grasp pair may still be infeasible in the current state because deformable objects exhibit severe self-occlusion, folding, and multi-layer stacking. Selection in Eq.~\ref{eq:select} and the verifier in the synthesis loop therefore share a set of closed-form predicates on the current cloth observation. \textbf{Occlusion-above} flags a candidate $u$ when a non-geodesic neighbor lies in the cylinder above it,
\begin{equation}
\exists\, v \in V \setminus \mathcal{G}_\delta(u): \quad \|\mathbf{p}_v^{xy}-\mathbf{p}_u^{xy}\| < r,\ \ z_v - z_u \in [\delta_z,\,h],
\label{eq:occlusion}
\end{equation}
where $\mathcal{G}_\delta(u)$ is the $\delta$-radius geodesic neighborhood of $u$ on the canonical mesh; masking $\mathcal{G}_\delta(u)$ separates self-curvature from true occluding overlap. \textbf{Layer separation} flags $u$ when the canonical-coordinate $\ell_\infty$ spread of its physical XY-neighborhood exceeds a threshold, remaining reliable on thin fabrics where simple $z$-difference heuristics fail. Additional predicates enforce reachability, anti-cross-arm safety, and surface exposure on the current 2-D convex hull, augmenting the candidate set when canonical landmarks fold inward. These predicates gate selection, terminate synthesis upon success, and define task success during evaluation; being closed-form functions of the observation, they keep the loop deterministic and label-noise-free.

\paragraph{Closed-loop synthesis.}
Grasp failure is among the most common failure modes for deformables, particularly under occlusion and self-contact, so trajectory synthesis cannot be a single-shot selection. Each grasp attempt in \syssyn{} couples topology-aware selection with verification: at every iteration, the system re-observes the cloth, updates the visible vertex set under the occlusion mask, re-evaluates Eq.~\ref{eq:select} under the latest feasibility constraints, executes the grasp while anchoring already-verified arms, and retries up to $T$ attempts (full pseudocode in Appendix~\ref{app:topogen-alg}).

\section{\sysreal{}: Sim-to-Real Protocol}
\label{sec:real}

Beyond rendering realism and standard domain randomization, we identify
two deformable-specific sim-to-real gaps and address each at the
appropriate layer: the cloth \emph{state distribution} that rigid-body
pose randomization cannot cover (\S\ref{sec:state-coverage}), and the
per-unit ISP noise of consumer RGBD cameras (\S\ref{sec:photoaug}).
We further apply conventional DR over robot and scene axes
(\S\ref{sec:dr-axes}). Demonstration synthesis (\S\ref{sec:gen}) and
in-simulation policy evaluation share the same environment, so
sim-to-sim closed-loop checks are available throughout development; the
generalization claims in this paper are evaluated against real-robot
deployment in \S\ref{sec:exp-main}.

\paragraph{Physics-driven cloth state randomization.}
\label{sec:state-coverage}
Deformable manipulation involves a high-dimensional state space of folds, wrinkles, and self-contact that simple pose randomization cannot capture. We therefore initialize cloth states using task-conditioned physics primitives (e.g., pin-and-release or random-fold-and-settle), allowing the simulator to generate physically plausible configurations through contact, gravity, and bending dynamics.

\paragraph{Sensor-distribution-aware photometric augmentation.}
\label{sec:photoaug}
Real-world deployment uses three RealSense D435i cameras (one overhead,
two wrist-mounted). We characterize their per-unit ISP behavior and
identify two sensor-internal failure modes---color bias across cameras
and gain-loop channel drift within a single camera---that conventional
pipelines often absorb into generic environmental randomization. We
instead apply photometric augmentation with ranges fitted to the
measured per-camera statistics; disabling it collapses real-world
success to $0\%$ on all five tasks, while enabling it restores the
rates reported in \S\ref{sec:exp-main}. Procedure and ranges are in
Appendix~\ref{app:sim-real-cal}.

\paragraph{Robot and scene domain randomization.}
\label{sec:dr-axes}
Conventional DR covers per-joint robot home-pose noise, table height, table texture, and lighting. Deformable object physical parameters, such as bending stiffness and friction, are matched to direct measurements of the corresponding real fabrics without task-specific tuning, enabled by \syssim{}'s physically meaningful parameterization (\S\ref{sec:sim}). All axes
are sampled per episode via a scrambled Sobol quasi-random sequence to
improve coverage under the 200-episode budget; full ranges in
Appendix~\ref{app:dr}.



\section{\sysasset{}: Extensible Asset Framework}
\label{sec:asset}

\sysasset{} is an extensible deformable-asset framework for sim-to-real research. It bundles approximately $2{,}000$ garment meshes imported from CLOTH3D~\citep{cloth3d} together with additional asset categories underrepresented in prior deformable datasets, most notably plastic bags. All assets share a unified triangle-mesh representation, enabling easy interchange across solvers and renderers.

A key feature of \sysasset{} is its measurement-backed physical parameterization. Unlike PBD-based simulators whose parameters, such as inter-particle distance, often lack direct real-world correspondence, \sysasset{} represents material properties as measurable physical quantities (mass per unit area, bending stiffness, stretch, friction). The physical-property values used in this paper are drawn from the large-scale fabric measurement library released by RGBench~\citep{hu2026RGBench}, which catalogs real fabrics characterized under standard textile-testing protocols (e.g., ASTM~D1388 cantilever bending~\citep{astm_d1388}, ASTM~D3107 stretch/extensibility~\citep{astm_d3107}). Each asset is bound to a fabric class entry from this library, so sim-to-real experiments use physically grounded settings rather than manually tuned simulator proxies.

To further scale asset collection, \sysasset{} integrates a single-image asset-generation pipeline based on EmbodiedGen~\citep{embodiedgen}. Unlike standard 3D generation pipelines that mainly recover geometry and appearance, our pipeline additionally assigns deformable physical properties to generated assets by sampling from the same RGBench-derived property library according to the inferred fabric category, optionally predicted by a VLM-based classifier. As a result, generated assets can be directly instantiated in \syssim{} with measurement-backed dynamics, avoiding generic defaults or manual tuning. Imported meshes and generated assets share the same physical-parameter interface, so the collection grows without retuning the simulator; release details are in Appendix~\ref{sec:asset-release}.

\section{Experiments}
\label{sec:exp}

\subsection{Setup}
\label{sec:exp-setup}
Our bimanual setup---two Piper 6-DOF arms with parallel-jaw grippers, one overhead and two wrist-mounted RealSense D435i cameras---is described in detail in \S\ref{sec:real}. We train $\pi_{0.5}$~\citep{pi05} with full fine-tuning on the 200 demonstrations generated by \syssyn{} per task. A DP3~\citep{dp3} point-cloud baseline trained on the same demonstrations fails on all five real tasks; the failure reproduces across the consumer RGBD sensors we tested, so we attribute it to depth-acquisition limits on matte-black grippers and silk/bag surfaces rather than to point-cloud methods in general (see \S\ref{sec:real} and Appendix~\ref{app:failures}).

\subsection{Simulator Reliability and Efficiency}
\label{sec:exp-sim}
We benchmark \syssim{} against two cloth simulators previously used for deformable data generation: Isaac~Sim with PhysX particle cloth (parameters inherited from DexGarmentLab~\citep{wan2025dexgarmentlab}) and the Newton implementation of VBD~\citep{chen2024vbd}, each under its officially recommended parameters. All three systems import the same garment asset and execute the same predefined dual-arm grasp-and-lift; per-step time is averaged over 200 physics steps after a 100-step warm-up at a $1/500$\,s timestep on the same hardware.
Figure~\ref{fig:sim-failures} shows representative failure frames.

Table~\ref{tab:sim-reliability} compares simulator reliability on repeated bimanual garment grasping. Isaac Sim fails to establish stable grasps, while Newton VBD frequently suffers from physically invalid contact failures such as penetration and explosion. In contrast, \syssim{} achieves stable and efficient simulation without these failure modes, resulting in substantially higher usable demonstration yield for large-scale data generation.

\begin{table}[h]
  \caption{\textbf{Reliability comparison on repeated bimanual garment grasping}. Penetration / explosion are physically invalid contact failures; per-step time is at fixed asset and timestep.}
  \label{tab:sim-reliability}
  \centering
  \footnotesize
  \setlength{\tabcolsep}{4pt}
  \begin{tabular}{lccccc}
    \toprule
    Simulator & Task success $\uparrow$ & Grasp success $\uparrow$ & Penetration $\downarrow$ & Explosion $\downarrow$ & Per-step time $\downarrow$ \\
    \midrule
    Isaac~Sim (PhysX particle)~\citep{wan2025dexgarmentlab} & 0.0\% & 0.0\% & 0.0\% & 0.0\% & 7.80 ms \\
    Newton VBD~\citep{chen2024vbd} & 0.0\% & 100.0\% & 77.5\% & 22.5\% & 10.38 ms \\
    \textbf{Ours} & \textbf{100.0\%} & \textbf{100.0\%} & \textbf{0.0\%} & \textbf{0.0\%} & \textbf{4.44 ms} \\
    \bottomrule
  \end{tabular}
\end{table}

\subsection{Trajectory Synthesis Quality and Replay Stability}
\label{sec:exp-gen}
We assess \syssyn{} on three axes: \textbf{Pass} (synthesis success rate over $n$ trajectories); \textbf{Replay} (a successful trajectory replayed $100\!\times$ from fresh simulator resets, isolating contact determinism); and a two-stage failure decomposition (\textbf{Lift fail} / \textbf{Stall fail}) defined in Appendix~\ref{app:exp-readout}. We ablate two structural components: \textbf{w/o \textsc{tas}} replaces $E$ in Eq.~(\ref{eq:select}) with $V \times V$ (uniform sampling over visible-and-reachable bimanual pairs); \textbf{w/o closed-loop} restricts Algorithm~\ref{alg:topogen} to a single attempt ($T=0$). For cross-method comparison we run SIM1's official released train--sample--filter pipeline~\citep{zhou2026sim1} on its T-shirt fold task under identical physics.

\begin{table}[h]
\centering
\footnotesize
\caption{\textbf{Trajectory synthesis: pass rate, replay stability, and two-stage failure decomposition.} \emph{Top}: ablation of \syssyn{}'s two structural components on T-shirt flatten ($n=300$, same simulator and task). \emph{Bottom}: cross-method comparison on T-shirt fold ($n=100$, matched physics).}
\label{tab:gen-comparison}
\begin{tabular*}{\linewidth}{@{\extracolsep{\fill}}lcccccc}
\toprule
Method & Task & $n$ & Pass (\%) & Replay $100\!\times$ & Lift fail (\%) & Stall fail (\%) \\
\midrule
\syssyn{}                  & T-shirt flatten & 300 & \textbf{89.7} & \textbf{100/100} & 10.3 & 0    \\
\quad w/o closed-loop      & T-shirt flatten & 300 & 65.0          & ---                                 & 35.0 & 0    \\
\quad w/o \textsc{tas}     & T-shirt flatten & 300 & 56.3          & ---                                 & 43.7 & 0    \\
\midrule
\syssyn{}                  & T-shirt fold    & 100 & \textbf{97.2} & \textbf{100/100} & 2.8  & 0    \\
SIM1~\citep{zhou2026sim1}  & T-shirt fold    & 100 & 24.0          & 13/100  & 43.0 & 32.0 \\
\bottomrule
\end{tabular*}
\end{table}

\textbf{Ablation (flatten).} Removing closed-loop iteration substantially reduces pass rate, and further removing \textsc{tas} causes an additional drop. Most failures are lift failures, indicating that both closed-loop retry and topology-aware pair selection are critical for reliable trajectory synthesis.

\textbf{Cross-method comparison (fold).} Under matched $n=100$, \syssyn{} substantially outperforms SIM1~\citep{zhou2026sim1} in T-shirt folding. SIM1 suffers from both generator-side and simulator-side failures: the learned trajectory sampler introduces unstable grasps and large trajectory noise, while the contact solver frequently fails under bimanual cloth interactions, leading to lift and stall failures respectively.

\textbf{Replay isolates the simulator-side cause.} To separate the two causes, we replay one successful trajectory $100\times$ from fresh simulator resets on each system. \syssyn{} achieves \textbf{100/100} replay success, while SIM1 reaches only \textbf{13/100}. Since the trajectory is fixed across replays, the gap isolates simulator-side contact-resolution determinism, ruling out generator noise.

\subsection{Main Sim-to-Real Results Across Five Tasks}
\label{sec:exp-main}
The five tasks are: \textbf{snack packaging} (insert a snack into a plastic bag, then lift), \textbf{garment folding} (T-shirt fold), \textbf{garment unfolding} (T-shirt flatten from a wrinkled start), \textbf{silk unfolding} (silk flatten from arbitrary draping), and \textbf{silk grasping} (bimanual thin-silk grasp). They span bag--object interaction, structured garment manipulation, and thin-fabric handling under large deformation. Snack packaging in particular extends the suite beyond cloth-only settings to plastic-bag manipulation, an open category in prior sim-to-real work.

\begin{table}[h]
  \caption{\textbf{Main sim-to-real result} ($n=23$ consecutive real-robot trials per task; 200 sim demos per task, zero-shot bimanual deployment). Per-task and pooled Wilson $95\%$ CIs are in Appendix Tab.~\ref{tab:main-ci}.}
  \label{tab:main-results}
  \centering
  \footnotesize
  \setlength{\tabcolsep}{6pt}
  \resizebox{\linewidth}{!}{%
  \begin{tabular}{lcccccc}
    \toprule
     & Snack packaging & Garment folding & Garment unfolding & Silk unfolding & Silk grasping & Average \\
    \midrule
    Real success (\%) & 86.96 & 91.30 & 82.61 & 95.65 & 100.00 & \textbf{91.30} \\
    \bottomrule
  \end{tabular}%
  }
\end{table}

\begin{figure}[!htbp]
  \centering
  \includegraphics[width=\linewidth]{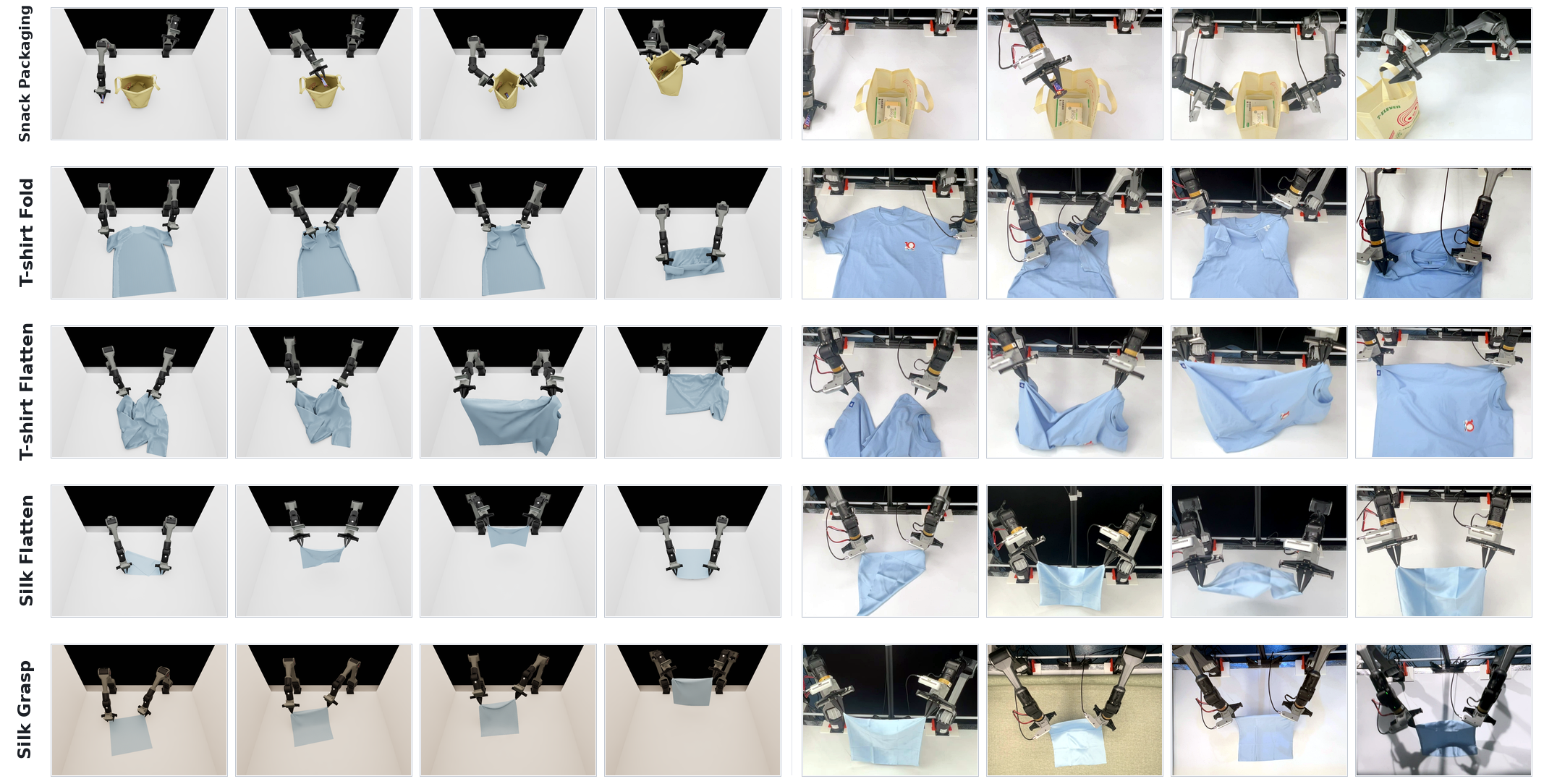}
  \caption{\textbf{Sim-to-real qualitative comparison across five deformable-manipulation tasks.} Within each row, the simulated rollout (left) and the zero-shot real-robot rollout (right) share the same checkpoint; one policy is trained per task on 200 sim demos, so checkpoints differ across rows.}
  \label{fig:main-sim-vs-real}
\end{figure}

Real-world success ranges from $82.61\%$ (garment unfolding) to $100\%$ (silk grasping), with a mean of $\mathbf{91.30\%}$ over $n=23$ consecutive trials per task (Wilson $95\%$ CIs in Appendix Tab.~\ref{tab:main-ci}). The two lowest cases---snack packaging ($86.96\%$) and garment unfolding ($82.61\%$)---each require recovery from wrinkled or arbitrarily-draped initial states, and their CIs overlap the higher-scoring tasks. The same $200$-demonstration budget transfers stably across all three task families. Additional setup details are provided in Appendix~\ref{app:exp-readout} and Appendix~\ref{app:dr}.

\subsection{Sample Efficiency and Generalization on Silk Grasping}
\label{sec:exp-generalization}
Silk grasping is a challenging generalization task due to weak texture, specular reflection, and sensitivity to grasp errors. We compare real-data policies trained on $25/50/100$ demonstrations with sim-data policies trained on $50/100/200$ demonstrations. Both pipelines share the same photometric augmentation; sim policies additionally receive scene-randomization axes (texture, lighting) unavailable to real-data collection (matched-augmentation discussion in Appendix~\ref{app:exp-readout}). Evaluation uses $n=23$ real-robot trials per cell under in-distribution scaling and OOD shifts in texture, lighting, and rotation.

\begin{figure}[t]
  \centering
  \begin{minipage}[t]{0.49\linewidth}
    \centering
    \includegraphics[width=\linewidth]{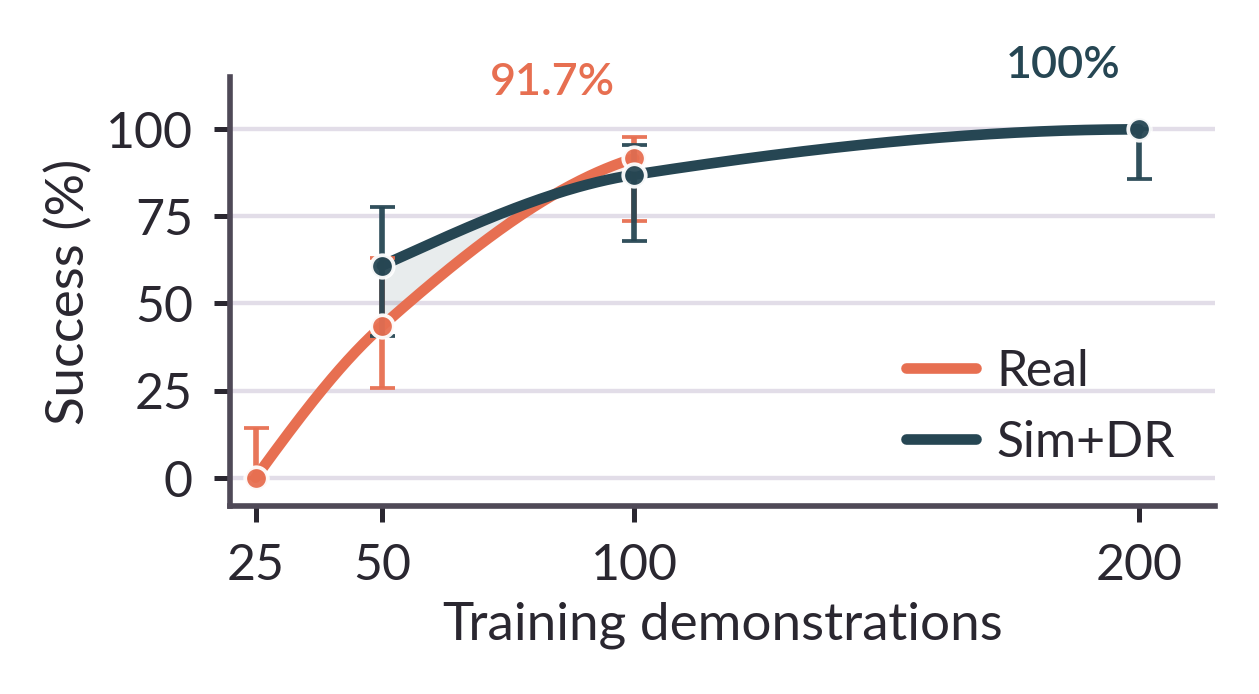}
  \end{minipage}\hfill
  \begin{minipage}[t]{0.49\linewidth}
    \centering
    \includegraphics[width=\linewidth]{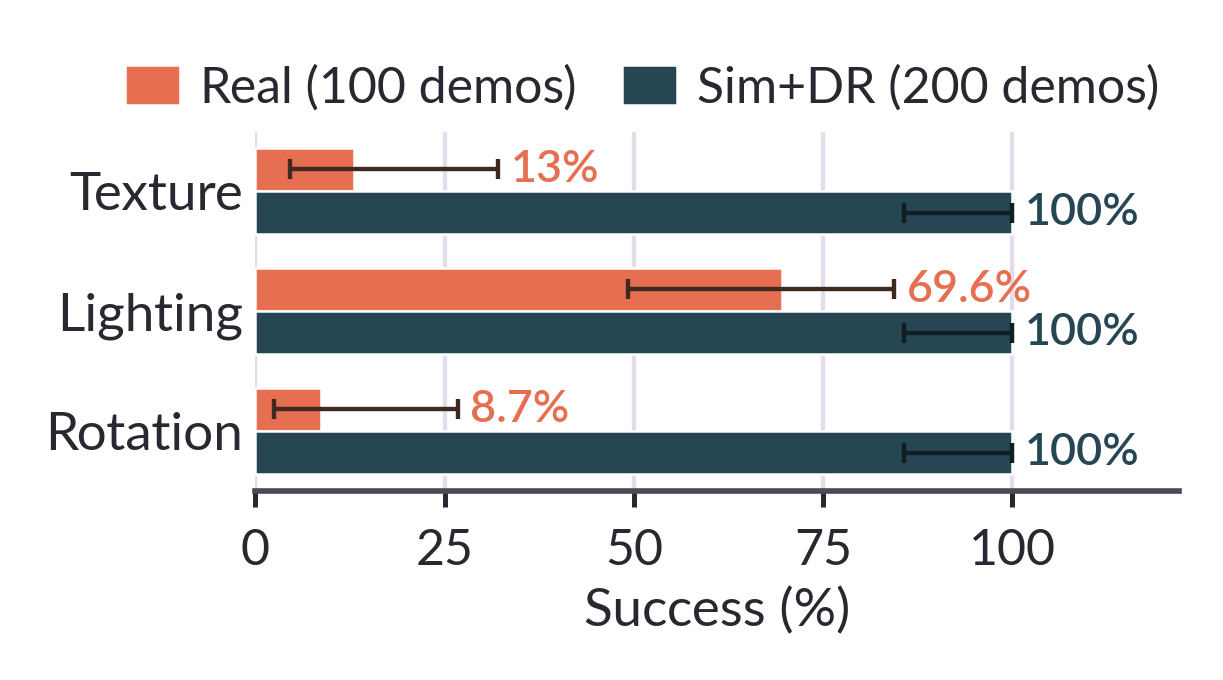}
  \end{minipage}
  \caption{\textbf{Silk grasping: real vs.\ sim. \textbf{(a)} In-distribution scaling.} \textbf{(b)} OOD under texture, lighting, and rotation shifts.}
  \label{fig:silk-real-vs-sim}
\end{figure}

\textbf{Sample efficiency (in-distribution scaling, panel a).} The sim-trained policy outperforms real-data training at small data scales and reaches $100\%$ at $n=200$; we read panel (a) as a scaling \emph{trend} rather than a per-budget significance test (Wilson 95\% CIs in Appendix~\ref{app:exp-readout}). The trend suggests that consistent synthesized trajectories are more sample-efficient than noisy teleoperated demonstrations, whose operator-specific variations require larger datasets to average out.

\textbf{Generalization under distribution shift (panel b).} On texture, lighting, and rotation shifts the sim-trained policy ($200$ demos + DR) scores $100\%$ uniformly versus $13.0\%$, $69.6\%$, and $8.7\%$ for real ($100$ demos)---all gaps significant under single-tail Fisher ($\alpha=0.05$). The lighting gain is partly a side effect of the photometric augmentation shared by both pipelines; texture and rotation gaps come from simulator-only DR axes (full CIs in Appendix~\ref{app:exp-readout}).

\paragraph{Cost efficiency.}
Reporting per-\emph{usable} rather than per-raw trajectory matters here: fast solvers that emit unstable rollouts incur post-hoc filtering whose discards inflate the true unit cost of training-ready data. Under matched server-cost assumptions ($8\times$RTX~4090), \syssyn{} reaches \textbf{2824} trajectories/day at \textbf{\$0.03} each---$4\times$ SIM1's throughput at $3\times$ lower unit cost, \textbf{two orders of magnitude cheaper} than real-robot collection (Appendix~\ref{app:cost-eff}).



\section{Conclusion}
\label{sec:conclusion}

We presented \sysname{}, which trains zero-shot RGB VLA policies on 200 sim demos per task to reach an average of $91\%$ real-world success ($>$80\% per task) across five deformable tasks---including plastic-bag manipulation---without teleoperation, by combining a measurement-backed simulator and asset framework (\syssim, \sysasset), a deterministic topology-aware synthesizer (\syssyn), and an ISP-aware sim-to-real protocol (\sysreal). On silk grasping, the sim-trained policy matches real-data training in-distribution and exceeds it under distribution shift, at two orders of magnitude lower cost (limitations in Appendix~\ref{sec:limitations}).

\begin{ack}
\end{ack}

\bibliographystyle{unsrt}
\bibliography{references}

\clearpage
\appendix
\section*{Appendix}
\addcontentsline{toc}{section}{Appendix}

\section{Domain Randomization Configuration}
\label{app:dr}

Domain randomization is configured per task. The silk-grasping task
receives the broadest DR --- including lighting, texture, and a
table-height axis --- because it serves as the centerpiece of the
visual-generalization study (\S\ref{sec:exp-generalization}). The
remaining four tasks use only cloth-pose, cloth-yaw, init-primitive,
and (where indicated) joint-noise randomization at task-appropriate
scales.

\paragraph{Per-task randomization ranges.}
\begin{table}[h]
  \centering\footnotesize
  \begin{tabular*}{\linewidth}{@{\extracolsep{\fill}}lccccc}
    \toprule
    Axis & Snack pkg. & G.\ fold & G.\ unfold & Silk unfold & Silk grasp \\
    \midrule
    Cloth XY offset           & $\pm 2$\,cm      & $\pm 5$\,cm      & --                  & $\pm 5$\,cm      & $\pm 10$\,cm \\
    Cloth yaw (Z)             & $\pm 5^{\circ}$  & $\pm 15^{\circ}$ & --                  & $\pm 10^{\circ}$ & $\pm 15^{\circ}$ \\
    Init primitive            & free-fall        & free-fall        & 2-pt hang-and-drop  & random fold     & free-fall \\
    Joint-noise scale         & $0.5\times$      & off              & off                 & $0.5\times$      & $0.5\times$ \\
    Table height              & --               & --               & --                  & --               & $-1.5 \to 0$\,cm \\
    Light exposure            & --               & --               & --                  & --               & $\pm 0.7$ stops \\
    Color temperature         & --               & --               & --                  & --               & $4000$--$6500$\,K \\
    Dome ambient              & --               & --               & --                  & --               & $0$--$0.15$ \\
    Table texture             & --               & --               & --                  & --               & $10{,}000$ var.$^{\dagger}$ \\
    Sampling                  & uniform          & uniform          & uniform             & uniform          & Sobol \\
    \bottomrule
  \end{tabular*}
  \caption{Per-task domain randomization ranges. ``--'' indicates the
  axis is disabled for that task. Joint-noise scale is relative to a
  task-conditional preset and applied symmetrically to both arms.
  Table-height randomization is asymmetric ($-1.5 \to 0$\,cm) because
  thin-fabric grasping is sensitive to downward calibration error but
  not to upward offset. $^{\dagger}$Texture set: $10{,}000$ training
  variants from the RoboTwin background-texture library
  \citep{robotwin}, with $1{,}000$ held-out variants reserved for the
  out-of-distribution evaluation in \S\ref{sec:exp-generalization}.}
  \label{tab:dr-per-task}
\end{table}

\paragraph{Per-task initialization protocol.}
The \emph{Init primitive} row of Tab.~\ref{tab:dr-per-task} encodes the
task-specific cloth-state generator that produces each episode's starting
configuration. Free-fall settle is used wherever the task naturally
operates from an arbitrary draped state (snack packaging, garment folding,
silk grasping). The two flattening tasks instead use a structured
generator that already determines the starting configuration's pose
family: the T-shirt (garment unfolding) is initialized by pinning two
key points and releasing under gravity, producing a wrinkled but
recognizable shape that matches how the task is presented downstream;
the silk (silk unfolding) is initialized via a random fold of two
diagonal corners. Because these generators already place the cloth in
a task-appropriate distribution, explicit cloth-XY-offset and yaw
randomization are unnecessary on top and the corresponding cells in
Tab.~\ref{tab:dr-per-task} read ``--''.

\paragraph{Sampling.}
For the silk-grasping task, where covering the joint randomization
space at the $200$-episode budget is critical, all axes are drawn per
episode from a scrambled Sobol quasi-random sequence. The other four
tasks use independent uniform sampling. At small data scales we
empirically find Sobol's better joint-axis coverage outweighs the
extra implementation cost.

\paragraph{Cloth physical parameters.}
Cloth physical parameters --- per-axis stretch and bend stiffness,
density, thickness, and gripper static/dynamic friction --- are taken
from a pre-built, \emph{industrially measured} fabric library: each
fabric class is characterized once under standard textile-testing
protocols (RGBench~\citep{hu2026RGBench}), and we assign every asset
the measured values of its matching fabric class. The parameters are
therefore measurement-backed rather than hand-tuned, yet deployment
requires \emph{no per-asset measurement and no per-task calibration}:
binding an asset to its fabric class is a one-time library lookup, and
the resulting values are neither randomized nor tuned per task. This
lookup is feasible only because the simulator's parameters are in
direct correspondence with physically meaningful quantities
(\S\ref{sec:sim}): each denotes a real, measurable property carried by
the library entry, rather than an opaque simulator proxy that must be
fitted by trial and error.

\paragraph{Table-height-randomization ablation (silk grasping).}
We isolate the contribution of the table-height axis on silk grasping
in simulation, holding all other axes fixed:

\begin{table}[h]
  \centering\footnotesize
  \begin{tabular}{lc}
    \toprule
    Configuration & Simulation success rate \\
    \midrule
    All randomization \emph{except} table height & $90\%$ \\
    All randomization \emph{including} table height & $97\%$ \\
    \bottomrule
  \end{tabular}
\end{table}

\section{Image Augmentation Pipeline}
\label{app:sim-real-cal}

\begin{figure}[h]
  \centering
  \includegraphics[width=\linewidth]{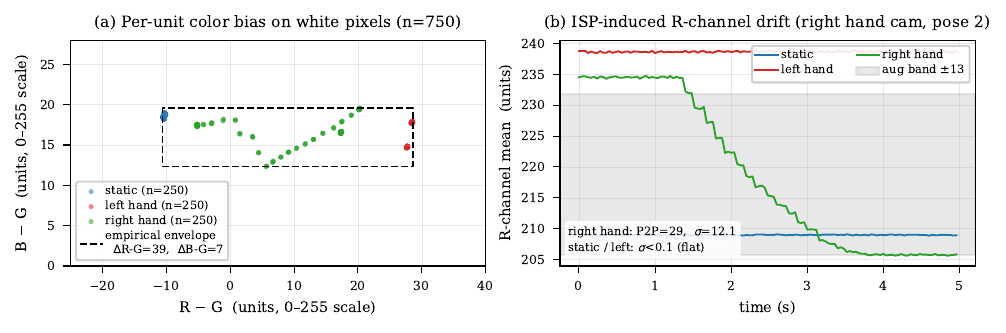}
  \caption{\textbf{Empirical D435i ISP characterization grounding the
  augmentation design.} (a) Per-unit color bias on white pixels across
  three cameras over two robot poses ($250$ frames each). The empirical
  bounding envelope spans $\Delta(R{-}G)=39$, $\Delta(B{-}G)=7$ units.
  (b) Frame-time R-channel drift on the third camera under pose 2:
  P2P~$=29$, $\sigma=12.1$, while the other two cameras are flat
  ($\sigma < 0.1$). Augmentation ranges in our pipeline (hue
  $\pm 0.15$ and per-channel jitter $\sigma=0.05 \approx \pm 13$
  units) are chosen against these empirical envelopes rather than
  literature defaults.}
  \label{fig:d435i-envelope}
\end{figure}

\paragraph{Empirical sensor characterization.}
We instrument all three Intel RealSense D435i units in the deployment
rig (one overhead, two wrist-mounted) under fixed manual exposure,
gain, and white-balance settings, and capture $125$ frames per camera
across two distinct robot poses ($750$ frames total). For each frame
we compute per-channel mean intensity on white pixels (all channels
$> 200/255$). Two systematic failure modes emerge
(Fig.~\ref{fig:d435i-envelope}). First, \emph{per-unit color bias}:
identical sensor settings yield camera-dependent color casts spanning
$\Delta(R{-}G) \approx 39$ units on a $0$--$255$ scale; two of three
cameras are pose-stable to within $\Delta < 0.2$ units across the two
poses, while the third drifts. Second, \emph{ISP gain-loop drift}:
the third camera's R-channel mean drifts P2P~$\approx 29$ units within
a single $5$-second window ($\sigma \approx 12$), with the drift
triggered by scene/angle changes and not removable by pre-deployment
calibration. Both effects are sensor-internal and require absorbing
on the simulation side rather than calibrating out at capture time.

A residual sim-real brightness offset of $7$--$10\%$ (simulator
brighter than capture) is also present and is absorbed by the same
photometric pipeline.

\paragraph{Training stack.}
Our policy training is built on the LeRobot stack~\citep{lerobot}, so
that data format, dataloader, training scripts, and augmentation
tooling all integrate with the open-source robotics community.

\paragraph{Augmentation parameters.}
The augmentation pipeline reuses LeRobot's photometric transforms,
with the brightness and hue sampling weights elevated and the hue
range expanded relative to its defaults to cover the D435i
cross-camera color bias.

\begin{table}[h]
  \centering\footnotesize
  \begin{tabular}{llll}
    \toprule
    Transform & Type & Range & Sampling weight \\
    \midrule
    Brightness  & ColorJitter      & $[0.75, 1.15]$            & $3.0$ \\
    Hue         & ColorJitter      & $[-0.15, 0.15]$           & $3.0$ \\
    Contrast    & ColorJitter      & $[0.80, 1.20]$            & $1.0$ \\
    Saturation  & ColorJitter      & $[0.50, 1.50]$            & $1.0$ \\
    Sharpness   & SharpnessJitter  & $[0.50, 1.50]$            & $1.0$ \\
    Affine      & RandomAffine     & $\pm 5^{\circ}$, $\pm 5\%$ trans. & $1.0$ \\
    \bottomrule
  \end{tabular}
  \caption{Per frame, a random subset of three transforms is drawn
  (with weights as sampling probabilities) and applied in random order.}
  \label{tab:aug-params}
\end{table}

\paragraph{Augmentation is necessary for transfer.}
We ablate the pipeline by training otherwise-identical policies with
all photometric augmentation disabled. Real-world success rates collapse
to $0\%$ uniformly across all five tasks:

\begin{table}[h]
  \centering\footnotesize
  \begin{tabular}{lcc}
    \toprule
    Task & w/ image augmentation & w/o image augmentation \\
    \midrule
    Snack packaging   & $86.96\%$  & $0\%$ \\
    Garment folding   & $91.30\%$  & $0\%$ \\
    Garment unfolding & $82.61\%$  & $0\%$ \\
    Silk unfolding    & $95.65\%$  & $0\%$ \\
    Silk grasping     & $100.00\%$ & $0\%$ \\
    \midrule
    Average           & $\mathbf{91.30\%}$ & $\mathbf{0\%}$ \\
    \bottomrule
  \end{tabular}
  \caption{Per-task real-world success rates with and without the
  photometric augmentation pipeline ($n=23$ trials per cell). The
  ``with augmentation'' column matches the main-result row of
  Tab.~\ref{tab:main-results}; the ``without'' column is the
  ablation, in which all five tasks collapse to $0\%$ uniformly.
  Image augmentation is a prerequisite for sim-to-real transfer, not
  a marginal improvement.}
  \label{tab:aug-ablation}
\end{table}

\section{Real-Robot Deployment}
\label{app:real}

\paragraph{Render-pipeline latency.}
RGB observations are produced by one Omniverse Kit frame update per
simulation tick (\texttt{kit\_frame\_updates=1}), so each rendered
image lags the underlying simulator state by one to two control steps
--- approximately $40$--$80$\,ms at the $25$\,Hz control rate. This
deliberately matches the natural capture-to-output pipeline latency of
the D435i RGB stream ($30$--$50$\,ms in our measurements), avoiding
the unrealistically instantaneous-render regime that a default
zero-lag rendering loop would impose on the training data. We do not
isolate the contribution of this setting with a dedicated ablation;
we report it for reproducibility.

\section{\syssyn{} Pseudocode}
\label{app:topogen-alg}

\begin{algorithm}[H]
  \caption{\syssyn{}: closed-loop topology-aware grasp synthesis.}
  \label{alg:topogen}
  \begin{algorithmic}[1]
    \Require topology graph $G$, $\textsc{Observe}(\cdot)$, attempt budget $T$
    \For{$t = 0, \ldots, T$}
      \State $\text{obs} \gets \textsc{Observe}()$;\quad $(l^{\star}, r^{\star}) \gets \textsc{TopoSelect}(G,\, \text{obs})$ \hfill\Comment{Eq.~(\ref{eq:select})}
      \State Execute grasp at $(l^{\star}, r^{\star})$, anchoring already-verified arms at $\text{obs}.q$
      \State $(l_{\text{ok}}, r_{\text{ok}}) \gets \textsc{Verify}(\textsc{Observe}())$
      \If{$l_{\text{ok}} \wedge r_{\text{ok}}$}\ \Return $\textsc{Execute}(\tau_{\text{post}})$ \EndIf
    \EndFor
    \State \Return \textsc{abort}
  \end{algorithmic}
\end{algorithm}

\section{Topology-Graph Instantiations Across Assets}
\label{app:topo-instantiations}

The same selection--verification loop (Algorithm~\ref{alg:topogen}) applies across all three asset categories. Only the canonical mesh and the graph $G=(V,E)$ change between assets; predicates, the closed-loop iteration, and the verifier are shared.

\begin{table}[h]
  \centering\footnotesize
  \begin{tabular}{lll}
    \toprule
    Asset class & Landmark set $V$ & Adjacency $E$ \\
    \midrule
    Cloth (rectangular)        & 4 corners                                                          & 4 boundary edges; diagonals excluded \\
    Garment (T-shirt)          & 7 landmarks (neck, $\text{hem}_{l/r}$, $\text{shoulder}_{l/r}$, $\text{sleeve}_{l/r}$) & task-feasible subset (e.g.\ hem $\!\leftrightarrow\!$ sleeve) \\
    Bag (with strap)           & 2 strap-tips (one per handle)                                      & $\text{strap-tip}_l \!\leftrightarrow\! \text{strap-tip}_r$ \\
    \bottomrule
  \end{tabular}
  \caption{Topology graph $G=(V,E)$ instantiated per asset class.}
  \label{tab:topo-instantiations}
\end{table}

\paragraph{Graph construction across asset classes.}
The construction of $G=(V,E)$ takes one of two paths depending on the asset's symmetry.

\textit{Rectangular cloth.} For symmetric assets, $G$ is extracted automatically from the canonical mesh. We identify the boundary edges of the mesh (edges incident to exactly one triangle), traverse them into a closed boundary loop, and select four farthest-from-centroid vertices as corners; midpoints of the four boundary segments between corners serve as edge-center landmarks. The adjacency edge set $E$ contains the four boundary-aligned corner pairs and excludes the two diagonal pairs, since the cloth folds rather than unfolds along the diagonal axis under fling. No human input is required.

\textit{Garments and bags.} For semantically structured assets, the landmark vertex indices are labeled once per asset class on the canonical mesh during asset preprocessing. T-shirt classes carry $\{\text{neck}, \text{hem}_{l/r}, \text{shoulder}_{l/r}, \text{sleeve}_{l/r}\}$ as the typical $V$; bag-strap classes carry two strap-tip landmarks (one per handle). The edge set $E$ encodes the task-feasible bimanual pairs for that asset class (e.g., $\text{hem} \leftrightarrow \text{sleeve}$ for T-shirt sleeve folding; $\text{strap-tip}_l \leftrightarrow \text{strap-tip}_r$ for bag bimanual lift). Once labeled, both $V$ and $E$ are reused across all instances of the asset class and across all rendered episodes, so per-trajectory annotation is never required.

Across the entire training corpus, the human input on $G$ amounts to a one-time labeling of $|V|$ vertex indices per asset class, typically a few minutes per new garment or bag class, and zero for symmetric assets where $G$ is auto-extracted.

\paragraph{Geometric feasibility set $\mathcal{F}(\text{obs}_t)$.}
The feasibility set in Eq.~\ref{eq:select} is the intersection of six closed-form predicates, evaluated on the current observation $\text{obs}_t = \{p_v\}_{v \in V}$.

\textit{Reachability, anti-cross-arm, and safety.} A candidate vertex must lie within the workspace of at least one arm; a candidate pair must remain on the same side of the inter-base axis so the arms do not cross, and must maintain a physical safety margin:
\begin{align}
\mathcal{F}_{\text{reach}}(u)\;&:\quad \min\bigl(\|p_u - b_l\|,\ \|p_u - b_r\|\bigr) \;\leq\; R_{\text{ws}}, \\
\mathcal{F}_{\text{cross}}(u,v)\;&:\quad \langle p_u - p_v,\ b_l - b_r \rangle \;\geq\; 0, \\
\mathcal{F}_{\text{safe}}(u,v)\;&:\quad \|p_u - p_v\| \;\geq\; d_{\min}.
\end{align}

\textit{Surface exposure.} A candidate must lie near the cloth's current 2-D outline, allowing observation-driven hull vertices to augment $V$ when canonical landmarks fold inward and become unreachable:
\begin{equation}
\mathcal{F}_{\text{surf}}(u)\;:\quad \min_{q \in \mathcal{H}_{xy}(\text{obs}_t)}\, \|p_u^{xy} - q\| \;<\; d_{\text{hull}},
\end{equation}
where $\mathcal{H}_{xy}(\text{obs}_t)$ is the set of vertices on the 2-D convex hull of the observed cloth.

\textit{Occlusion-above.} Restating Eq.~\ref{eq:occlusion} as a feasibility predicate, $u$ is feasible only when no non-geodesic neighbor occupies the cylinder above it:
\begin{equation}
\mathcal{F}_{\text{occ}}(u)\;:\quad \neg\,\exists\, v \in V \setminus \mathcal{G}_\delta(u):\ \|p_v^{xy} - p_u^{xy}\| < r,\ \ z_v - z_u \in [\delta_z, h].
\end{equation}

\textit{Layer separation.} The canonical-coordinate $\ell_\infty$ spread of $u$'s physical XY-neighborhood $\mathcal{N}(u) = \{v: \|p_v^{xy} - p_u^{xy}\| < r\}$ must remain below a threshold:
\begin{equation}
\mathcal{F}_{\text{layer}}(u)\;:\quad \max_{i \in \{x, y\}} \Bigl(\max_{v \in \mathcal{N}(u)} c_i(v) \;-\; \min_{v \in \mathcal{N}(u)} c_i(v)\Bigr) \;\leq\; \tau_{\text{spread}},
\end{equation}
where $c(\cdot)$ denotes canonical-mesh position. Vertices physically close in 3-D but topologically far on the canonical mesh indicate multi-layer overlap.

\textit{Composition.} The full feasibility set is the intersection of these predicates, evaluated at both $u$ and $v$ where applicable:
\begin{equation}
\begin{aligned}
\mathcal{F}(\text{obs}_t) \;=\; \Bigl\{ (u, v) \in V \!\times\! V :\ &\mathcal{F}_{\text{cross}}(u,v) \,\wedge\, \mathcal{F}_{\text{safe}}(u,v) \\
&{} \wedge \bigwedge_{x \in \{u, v\}}\!\bigl( \mathcal{F}_{\text{reach}}(x) \,\wedge\, \mathcal{F}_{\text{surf}}(x) \,\wedge\, \mathcal{F}_{\text{occ}}(x) \,\wedge\, \mathcal{F}_{\text{layer}}(x)\bigr) \Bigr\}.
\end{aligned}
\end{equation}

\section{Multi-Backend Planner and Inverse Kinematics}
\label{app:planner-ik-backends}

The trajectory synthesizer is decoupled from any single planning or kinematics implementation through a unified \texttt{PlannerManager} interface and a \texttt{BaseKinematics} abstraction. Each backend is swappable at configuration time, and the synthesis loop never references a backend directly. This decoupling supports two practical needs: hardware-portability, since CPU-only deployments can fall back to analytical IK while GPU-equipped servers leverage batched IK, and faithful cross-platform reproduction of the synthesis pipeline without re-implementing higher-level logic.

\begin{table}[h]
  \centering
  \footnotesize
  \setlength{\tabcolsep}{5pt}
  \begin{tabular}{llll}
    \toprule
    Layer & Backend & Implementation & Notes \\
    \midrule
    \multirow{4}{*}{Motion planning}
      & TOPP-RA      & CPU, deterministic       & Default; time-optimal parameterization \\
      & cuRobo       & GPU, batched             & High-throughput data generation \\
      & Min-jerk     & CPU, analytical          & Reference baseline \\
      & Adaptive     & Hybrid                   & Per-segment backend selection \\
    \midrule
    \multirow{4}{*}{Inverse kinematics}
      & Pinocchio    & Analytical Jacobian, CPU & Default \\
      & cuRobo       & GPU, batched             & Pairs with cuRobo planner \\
      & RTB          & Robotics Toolbox         & Cross-validation \\
      & TRAC-IK      & Constrained quadratic    & Singular-configuration fallback \\
    \bottomrule
  \end{tabular}
  \caption{\textbf{Planner and IK backends sharing a unified interface.} The synthesizer is invariant to backend choice; selection is configuration-driven. The pipeline used throughout this paper combines TOPP-RA + Pinocchio as the default; cuRobo is enabled when batch IK throughput becomes the bottleneck.}
  \label{tab:backends}
\end{table}

\paragraph{Multi-stage skill composition.}
Skills with qualitatively different motion phases such as bimanual fling, drag, or contact-critical grasp cannot be expressed under a single fixed-parameter motion plan. The synthesizer therefore exposes per-stage specifications: each stage of a skill carries its own velocity scale, its own per-axis pose-constraint vector (in either world or end-effector frame), and its own goal type (Cartesian pose or joint configuration). A configurable via-point is inserted before each grasp to relax IK reachability without altering the camera viewpoint, and IK solutions propagate as warm-starts across consecutive stages to minimize wrist rotation.

The bimanual fling is the canonical example. It chains a slow lift to the fling start pose, a high-velocity forward swing that injects momentum into the cloth, a mid-low transition that absorbs dynamic forces, a low-$z$ drag with tightly locked gripper orientation that keeps the fabric flat against the table, and a back-amplitude return that resets the configuration. Each stage carries its own velocity scale and constraint vector, specified once at task-config time and consumed by the planner without solver-side intervention; no fixed-parameter pipeline could produce this trajectory uniformly.

\paragraph{Behavioral comparison with human teleoperation.}
As a post-hoc behavioral check, we compare \syssyn{}-generated trajectories with $100$ human-teleoperated episodes of the same task on the same hardware. The comparison is descriptive only: it asks whether synthesized trajectories sit within the same family of motion profiles as humans, not whether they are derived from human data. \syssyn{} does not consume any teleoperation in either training or planning.

We fit five candidate velocity-profile models to the normalized velocity curve of each human episode (Table~\ref{tab:profile-fit}). An asymmetric bell explains human motion best, with peak velocity at $\tau\!\approx\!0.35$ rather than the symmetric $\tau\!=\!0.5$ that min-jerk assumes. The synthesizer's TOPP-RA backend emits symmetric trapezoidal profiles; the median-MSE gap from symmetric trapezoidal to the best-fit asymmetric bell ($0.081$ vs.\ $0.007$) is roughly an order of magnitude smaller than gaps induced by initial-pose distribution mismatch and gripper-close timing. The synthesizer's profile family is therefore in-distribution with human motion behavior, with the residual shape difference dominated by other sim--real factors.

\begin{table}[h]
  \centering\footnotesize
  \begin{tabular}{lcc}
    \toprule
    Profile model & MSE (median) & Best-fit rate \\
    \midrule
    Asymmetric bell        & $\mathbf{0.007}$ & $\mathbf{97\%}$ \\
    Asymmetric trapezoidal & $0.056$ & $1.0\%$ \\
    Symmetric trapezoidal  & $0.081$ & $0.0\%$ \\
    Gaussian               & $0.088$ & $1.5\%$ \\
    Min-jerk               & $0.096$ & $0.5\%$ \\
    \bottomrule
  \end{tabular}
  \caption{Profile fit to $100$ human-teleoperated episodes; \syssyn{} does not consume teleoperation in any pipeline stage.}
  \label{tab:profile-fit}
\end{table}

\section{Failure Cases}
\label{app:failures}

\begin{figure}[h]
  \centering
  \includegraphics[width=0.78\linewidth]{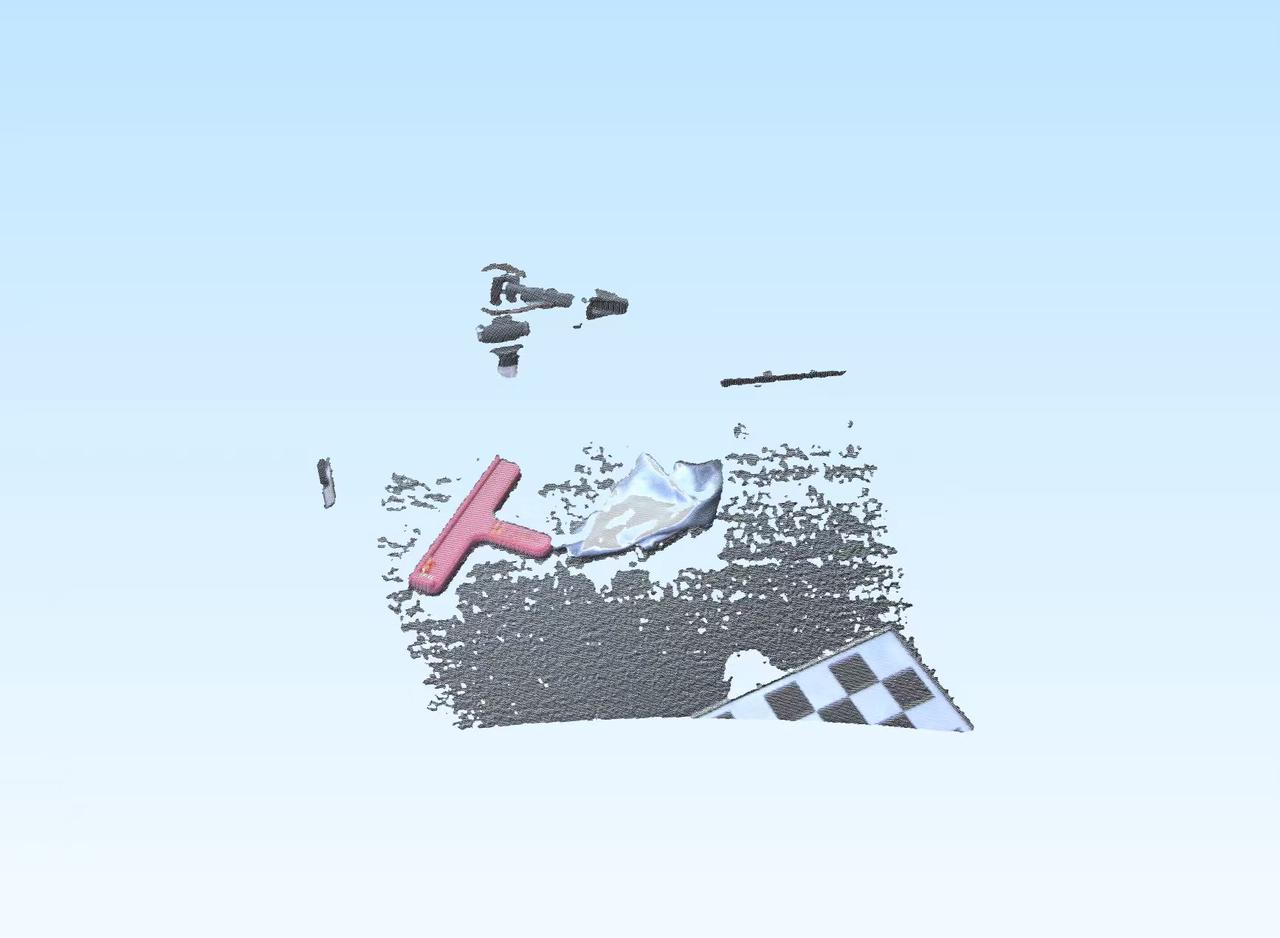}
  \caption{\textbf{Point-cloud coverage failure on our real-world scene
  (best-case scan).} A high-fidelity offline reconstruction --- the upper
  bound on what any depth-based pipeline can extract on this scene ---
  already exhibits large holes on both the table surface and the robot
  arm, collapses the gripper to a few isolated points, and recovers
  virtually no points on the silk fabric. The real-time D435i depth
  stream consumed by DP3 at deployment is strictly worse, which directly
  drives the universal failure of the point-cloud baseline across all
  five tasks.}
  \label{fig:dp3-failure}
\end{figure}

\paragraph{Point-cloud baseline (DP3) failure mode.}
We trained 3D Diffusion Policy (DP3)~\citep{dp3} on the same
\syssyn{}-generated demonstrations and evaluated on the same real
hardware. \textbf{DP3 fails across all five tasks.} The dominant cause
is fundamental coverage degradation of the point-cloud modality on our
scene. Fig.~\ref{fig:dp3-failure} shows a reconstruction captured by
our highest-fidelity scanning setup --- an upper bound on what any
depth-based pipeline can extract on this scene --- and even at this
quality, large portions of the table surface and the robot arm are
missing, the gripper collapses to a handful of sparse points, and the
silk fabric returns almost no signal. The real-time D435i depth stream
that DP3 actually consumes at deployment is strictly noisier and sparser
than this offline scan, so the policy never sees a reliable geometric
signature for either the manipulator or the cloth and loses track of
both before reaching the grasp.

\section{Extended Experiment Readout}
\label{app:exp-readout}

\paragraph{Failure-mode definitions (Tab.~\ref{tab:gen-comparison}).}
\textbf{Lift fail} counts trajectories where the gripper closes but does not lift the cloth past the height threshold---the cloth was not actually grasped. \textbf{Stall fail} counts trajectories where the cloth is lifted but its dynamic motion stalls during the subsequent fling/fold phase: contact-resolution interference (any of contact-force balancing, self-collision, or cloth--rigid collision) blocks the cloth from following the planned trajectory, leaving it in a wrong terminal configuration. This is a simulator-side breakdown rather than a planning failure.

\paragraph{Simulator metrics and failure interpretation.}
Section~\ref{sec:exp-sim} reports five metrics. \textbf{Task success} measures end-to-end completion. \textbf{Grasp success} measures whether the garment is successfully grasped and lifted before the release phase. \textbf{Penetration} measures cloth--rigid interpenetration or invalid post-contact attachment that makes the rollout physically implausible. \textbf{Explosion} measures catastrophic simulator instability with visually implausible cloth motion. For \textbf{per-step time}, we import the same asset into all simulators, fix the physics timestep to $1/500$\,s, discard the first 100 warm-up steps, and average the next 200 pure physics steps over repeated runs.

The detailed readout is consistent with the summary in the main text. Our simulator avoids the two failure modes that are most damaging for downstream data generation: visible cloth--rigid penetration and catastrophic instability. By contrast, the VBD-based baseline can nominally lift the garment but frequently exhibits penetration and explosive post-contact behavior, so apparently correct planned trajectories do not remain executable after replay: 77.5\% of trials exhibit penetration, 22.5\% end in explosion, and end-to-end task success drops to 0.0\%, while its per-step time rises to 10.38\,ms. Isaac~Sim cloth fails differently. It typically cannot establish or maintain a stable grasp at all, which results in 0.0\% grasp success, 0.0\% task success, and a slower per-step time of 7.80\,ms. At data-collection scale, both patterns produce large numbers of unusable demonstrations.

\paragraph{Replay protocol and trajectory-synthesis metrics.}
At the trajectory level, we start from a single successful garment-folding trajectory collected in our simulator, then reset the simulator and replay the generated trajectory under the same task definition to test whether the grasp and subsequent execution remain valid. This replay-based protocol directly validates the determinism claims made in Section~\ref{sec:gen}: the question is not whether a trajectory can succeed once, but whether it remains executable across repeated simulator resets.

Table~\ref{tab:gen-comparison} is organized along three axes. \textbf{Pass} is the success rate over $n$ independently synthesized trajectories. \textbf{Replay} is the success rate of a single successful trajectory replayed $100\!\times$ in our simulator, isolating simulator-side determinism under contact. \textbf{Lift fail / Stall fail} decompose the failure mass into mutually exclusive modes (defined above)---gripper closes but cloth is not lifted past the height threshold; cloth is lifted but its dynamic motion stalls during the fling/fold phase leaving the wrong terminal configuration---and sum to $1-\text{Pass}$ up to rounding. We ablate the two structural components introduced in Section~\ref{sec:gen}: removing the topology graph (\textbf{w/o \textsc{tas}}) replaces $E$ in Eq.~(\ref{eq:select}) with $V \times V$, sampling bimanual pairs uniformly from the visible-and-reachable subset; removing the closed-loop iteration (\textbf{w/o closed-loop}) restricts Algorithm~\ref{alg:topogen} to a single attempt ($T=0$). We further compare against SIM1's \emph{official released implementation and trajectory sampler} under its default configuration (not a reimplementation), evaluated on its original T-shirt fold task. The $100\!\times$ replay is conducted in SIM1's own released simulator, so the $13/100$ result reflects that solver's contact determinism under a fixed successful trajectory, independent of our pipeline.

\paragraph{Detailed synthesis readout.}
On T-shirt flatten, removing the closed-loop iteration drops \syssyn{}'s pass rate from 89.7\% to 65.0\%, isolating the closed-loop loop at $+24.7$\,pp; further removing the topology graph drops the rate to 56.3\%, isolating \textsc{tas} at an additional $+8.7$\,pp on top of single-attempt synthesis. The latter ablation samples bimanual pairs uniformly from the visible-and-reachable subset, so the $1/3$ probability of selecting a diagonal pair---which fails under fling because the cloth folds along the diagonal axis rather than unfolding---accounts for part of the remaining gap. On T-shirt fold under matched conditions with~\citep{zhou2026sim1}, \syssyn{} reaches 97.2\% versus 24.0\%, a 73-point same-task same-simulator gap.

A single successful trajectory replayed $100\!\times$ yields $100/100$ for \syssyn{} and only $13/100$ for~\citep{zhou2026sim1}. Because replay is independent of both the synthesis method and the task, this 87-point gap reflects contact-rich simulator determinism alone: trajectories produced by a learned sampler sit at the boundary of stable contact regimes, where small numerical perturbations cascade into task failure, whereas deterministic synthesis stays in the regime's interior.

\syssyn{}'s failures concentrate exclusively in the Lift-fail mode---10.3\% on flatten and 2.8\% on fold---the same mode that dominates~\citep{zhou2026sim1}'s failures at 43\%. The same-mode reduction is $4\!\times$ on flatten and $15\!\times$ on fold, which isolates the value of the closed-loop mechanism in the slip-prone grasp regime. \syssyn{} also produces zero Stall-fail on either task versus 32\% for~\citep{zhou2026sim1}, which is consistent with \textsc{tas} preventing structurally invalid grasps from cascading into wrong terminal configurations.

\paragraph{Main results: per-task Wilson 95\% CIs (Tab.~\ref{tab:main-results}).}
With $n=23$ trials per task and $N=115$ pooled trials, the per-task and pooled Wilson $95\%$ confidence intervals corresponding to the main sim-to-real result are:

\begin{table}[h]
  \centering\footnotesize
  \setlength{\tabcolsep}{6pt}
  \begin{tabular}{lcc}
    \toprule
    Task & Success rate (\%) & Wilson $95\%$ CI \\
    \midrule
    Snack packaging   & 86.96  & $[67.9,\ 95.5]$ \\
    Garment folding   & 91.30  & $[73.2,\ 97.6]$ \\
    Garment unfolding & 82.61  & $[62.9,\ 93.0]$ \\
    Silk unfolding    & 95.65  & $[79.0,\ 99.2]$ \\
    Silk grasping     & 100.00 & $[85.7,\ 100.0]$ \\
    \midrule
    Average ($N{=}115$) & \textbf{91.30} & $\mathbf{[84.7,\ 95.2]}$ \\
    \bottomrule
  \end{tabular}
  \caption{Per-task and pooled Wilson $95\%$ confidence intervals for the main sim-to-real result (Tab.~\ref{tab:main-results}); $n=23$ consecutive real-robot trials per task.}
  \label{tab:main-ci}
\end{table}

\paragraph{Silk grasping: matched-augmentation comparison and CIs (\S\ref{sec:exp-generalization}).}
The OOD comparison isolates the data-source axis (sim vs.\ real) at matched 2D image augmentation. Both pipelines apply identical photometric augmentation (color jitter, brightness/contrast, photometric ranges fitted to per-camera ISP statistics, \S\ref{sec:photoaug}); the only deliberate asymmetry is that the sim policies additionally receive appearance and lighting randomization that requires programmatic scene control and is therefore unavailable to real-data collection. Simulator-only DR is documented as an additional axis on the sim side rather than absorbed into the matched augmentation.

With $n=23$ trials per cell the Wilson 95\% confidence intervals are wide ($\pm{\sim}19$\,pp at the mid-range $50\%$ probability), so the in-distribution scaling result (panel a) is read as a trend rather than a per-budget significance test. For OOD (panel b), the sim-trained policy ($200$ demos + DR) scores $100\%$ uniformly with Wilson 95\% CI $[85.7, 100.0]$ versus $13.0\%$ ($[4.5, 32.1]$), $69.6\%$ ($[49.1, 84.4]$), and $8.7\%$ ($[2.4, 26.8]$) for real ($100$ demos) on texture, lighting, and rotation respectively; all gaps remain significant under a single-tail Fisher test ($\alpha=0.05$).

\section{Cost Efficiency}
\label{app:cost-eff}

\paragraph{Methodology (Table~\ref{tab:collection-cost}).}
``Single-trajectory time'' denotes the wall-clock time to obtain one usable trajectory before dataset serialization; for our pipeline this includes planning, physics execution, and Isaac rendering. ``Throughput'' denotes usable trajectories collected per day under the reported hardware setup. ``Unit cost'' denotes amortized cost per usable trajectory. ``Relative cost'' is normalized to the real-robot collection cost. Our throughput and unit-cost entries are computed from a measured $244.8$\,s per usable trajectory under the same $8\times$RTX~4090 server-cost assumption used in SIM1. SIM1 and real-robot numbers follow the cost-efficiency analysis reported in SIM1~\citep{zhou2026sim1}; the real-robot single-trajectory time is derived from $104$ trajectories collected over an $8$-hour workday.

\begin{table}[h]
  \caption{\textbf{Per-trajectory collection efficiency.} All quantities are per \emph{usable} trajectory under matched server-cost assumptions ($8\times$RTX~4090).}
  \label{tab:collection-cost}
  \centering
  \footnotesize
  \begin{tabular*}{\linewidth}{@{\extracolsep{\fill}}lcccc}
    \toprule
    Method & \makecell{Single-Trajectory\\Time (min/traj.) $\downarrow$} & \makecell{Throughput\\(usable traj./day) $\uparrow$} & \makecell{Unit Cost\\(\$/usable traj.) $\downarrow$} & \makecell{Relative\\Cost $\downarrow$} \\
    \midrule
    Real Robot    & 4.6          & 104           & 2.71          & 1.0$\times$ \\
    SIM1          & 16.2         & 710           & 0.10          & 0.037$\times$ \\
    \textbf{Ours} & \textbf{4.1} & \textbf{2824} & \textbf{0.03} & \textbf{0.011$\times$} \\
    \bottomrule
  \end{tabular*}
\end{table}

Beyond transfer performance, we also evaluate whether the same pipeline is efficient enough to serve as a practical data-generation engine. We compare our method against SIM1~\citep{zhou2026sim1} and direct real-robot collection using four metrics: single-trajectory time, throughput, unit cost, and relative cost. All quantities are reported per usable trajectory, so the comparison reflects the cost of data that can actually be retained for policy training rather than the cost of raw rollouts with mixed quality.

Table~\ref{tab:collection-cost} shows that the proposed pipeline is not only effective, but also substantially more scalable than prior alternatives. Under the same server-cost assumption used in SIM1, our measured single-trajectory time is $4.1$ minutes per usable trajectory, compared with $16.2$ minutes for SIM1 and $4.6$ minutes for direct real-robot collection. More importantly, because our trajectories can be generated in parallel with high yield, the system produces $2824$ usable trajectories per day, compared with $710$ for SIM1 and $104$ for direct real-world collection.

The cost advantage is equally clear. Our unit cost is \$$0.03$ per usable trajectory, compared with \$$0.10$ for SIM1 and \$$2.71$ for direct real-robot collection. In normalized form, this corresponds to a relative cost of $0.011\times$ the real-robot baseline, versus $0.037\times$ for SIM1. Under the same accounting rule, our method therefore delivers roughly four times the usable-trajectory throughput of SIM1 while reducing the cost per usable trajectory by more than a factor of three, and remains nearly two orders of magnitude cheaper than collecting the data directly on hardware.

The key point is that the comparison is not about raw simulator speed alone. A simulator can run quickly while still producing unstable or unusable data that must later be discarded. The relevant quantity is therefore the cost of producing trajectories that survive quality control and can enter the final training set. Under that definition, our pipeline combines low single-trajectory latency, high parallel throughput, and high usable-data yield, making it well suited for scalable deformable-data generation in practice.

\section{Training Details}
\label{app:training}

\paragraph{Policy backbone.}
We use a $\pi_{0.5}$-class vision--language--action (VLA) policy. The model consumes RGB observations from one fixed-scene camera and two wrist-mounted cameras together with the bimanual joint state, and predicts a chunk of $50$ joint-action steps. We perform full-parameter fine-tuning from the publicly released $\pi_{0.5}$ base checkpoint; the visual encoder is not frozen.

\paragraph{Training recipe.}
Each policy is trained for $30{,}000$ optimizer steps on $200$ \syssyn{}-generated demonstrations per task. The effective batch size is $32$ ($8$ per GPU $\times\,4$ GPUs). Optimization uses AdamW with cosine learning-rate decay; the peak learning rate is $2.5\!\times\!10^{-5}$ after a $1{,}000$-step linear warmup. We use mixed-precision bfloat16 with gradient checkpointing throughout, and quantile-based normalization on both state and action streams. Hardware is $4\!\times\!\text{NVIDIA RTX 4090}$ (48\,GB each).

\paragraph{Data format.}
Demonstrations are stored in the LeRobot v3 schema (Parquet for state and action streams, MP4 for visual observations). Physics simulation runs at $500$\,Hz; observations and actions are sub-sampled to the policy's $25$\,Hz control cadence, and the recorded action chunk length matches the policy's prediction horizon.

\paragraph{Visual augmentation.}
The photometric augmentation parameters and the ablation showing that augmentation is required for sim-to-real transfer are documented in Appendix~\ref{app:sim-real-cal}.

\section{Limitations and Future Work}
\label{sec:limitations}

\paragraph{Strict comparison.}
Our most rigorous head-to-head comparison with real-robot training is on silk grasping (\S\ref{sec:exp-generalization}); broader cross-task comparisons require additional real-robot data collection and are left to future work.

\paragraph{Release.}
\label{sec:asset-release}
We plan to make our pipelines (trajectory synthesis, asset generation, and the sim-to-real training stack) and a representative subset of deformable assets available to the community. A usable version of the \syssim{} simulator will likewise be made available, while the full industrial-grade collection is available under licensed access, reflecting the cost of acquiring high-quality industrial deformable measurements.

\paragraph{Benchmark.}
A comprehensive benchmark with standardized task taxonomies, evaluation protocols, and policy leaderboards---built on the simulation infrastructure presented here---is in preparation as a separate contribution.


\end{document}